%% file: camera_ready.tex
\documentclass[runningheads]{llncs}
\usepackage{graphicx}
\usepackage{amsmath,amssymb} % define this before the line numbering.
\usepackage{color}
\usepackage{booktabs}
\usepackage{multirow}
\usepackage{subfig}

\newcommand{\mixed}[1]{\texttt{Mixed\_#1}}

\begin{document}

\title{Actor-Centric Relation Network}
\titlerunning{Actor-Centric Relation Network}

\authorrunning{C.~Sun et al.}

\author{Chen Sun \and Abhinav Shrivastava \and Carl Vondrick \and\\Kevin Murphy \and Rahul Sukthankar \and Cordelia Schmid}
\institute{Google Research}

\maketitle

\begin{abstract}
Current state-of-the-art approaches for spatio-temporal action localization rely on detections at the frame level and model temporal context with 3D ConvNets. Here, we go one step further and model spatio-temporal relations to capture the interactions between human actors, relevant objects and scene elements essential to differentiate similar human actions. Our approach is weakly supervised and mines the relevant elements automatically with an actor-centric relational network (ACRN). ACRN computes and accumulates pair-wise relation information from actor and global scene features, and generates relation features for action classification. It is implemented as neural networks and can be trained jointly with an existing action detection system. We show that ACRN outperforms alternative approaches which capture relation information, and that the proposed framework improves upon the state-of-the-art performance on JHMDB and AVA. A visualization of the learned relation features confirms that our approach is able to attend to the relevant relations for each action.

\end{abstract}

\input{intro}

\input{related}

\input{approach}

\input{experiment}

\section{Conclusion}
This paper presents a novel approach to automatically determine relevant spatio-temporal elements characterizing human actions in video. Experimental results for spatio-temporal action localization demonstrate a clear gain and visualizations show that the mined elements are indeed relevant. Future work includes a description of an actor by more than one feature, i.e., a number of features representing different human body parts. This will allow to model relations not only with an actor, but also the relations to relevant human parts. Another line of work could be to look at higher-order spatio-temporal relations.

\subsubsection{Acknowledgement:} We thank Chunhui Gu, David Ross and Jitendra Malik for discussion and comments.

\bibliographystyle{splncs04}
\bibliography{egbib}
\end{document}

%% file: intro.tex
\section{Introduction}

Robust human action understanding will have a large impact in applications across robotics, security, and health. However, despite significant progress in visual recognition for objects and scenes~\cite{he2016resnet,krizhevsky2012imagenet,russakovsky2015imagenet,zhou2014learning}, performance on action recognition remains relatively low. Now that we have large, diverse, and realistic datasets such as AVA~\cite{ava_cvpr18}, SLAC~\cite{zhao2017slac}, and Charades~\cite{sigurdsson2016hollywood}, why has action recognition performance not caught up?

\begin{figure}[t]
    \centering
    \includegraphics[width=0.95\linewidth]{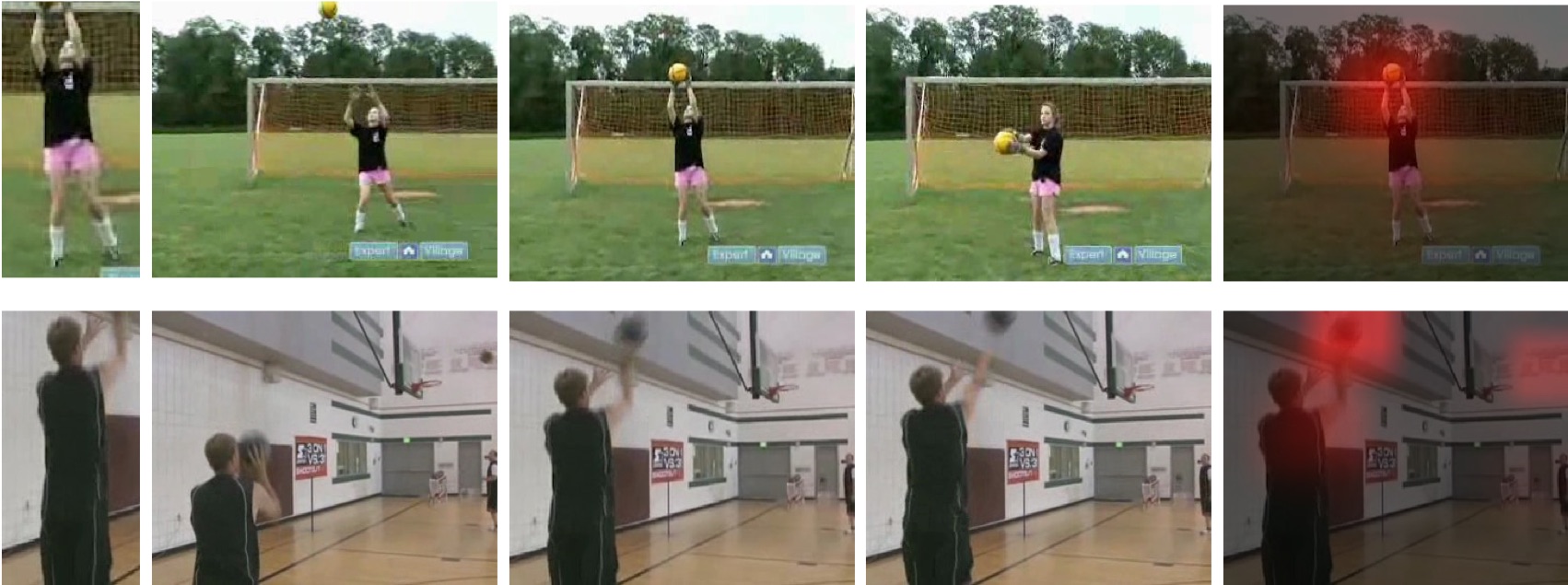}
    \caption{Action detection is challenging even for humans without relation reasoning from the context. Only by extracting the relationship between the actor and the object (ball), and understanding how this relationship evolves over time, can one tell that the first action is catching a ball, while the second action is shooting a ball. The last column visualizes the relational heat maps generated by our algorithm.
    }
    \label{fig:teaser}
    \vspace{-1em}
\end{figure}

Models for spatio-temporal action localization from the last few years have been mainly based on architectures for recognizing objects~\cite{gkioxari2017interactnet,peng2016multi,weinzaepfel2016}, building on the success of R-CNN style architectures \cite{girshick2015fast,girshick2014rich,ren2015faster}. However, unlike objects which can be identified solely by their visual appearance, in many cases actions can not be identified by the visual appearance of actors alone. Rather, action recognition often requires reasoning about the actor's relationship with objects and other actors, both spatially and temporally. To make this point, Figure~\ref{fig:teaser} shows two actors performing different actions. Even for humans, by just looking at the cropped boxes, it is difficult to tell what actions are being performed. It is from the actors' interactions with a ball in the scene that we can tell that these are sports actions, and only by temporal reasoning of the relative positions, can we tell that the first actor is catching a ball and the second is shooting a ball.

Although the basic idea of exploiting context for action recognition is not new, earlier works~\cite{dai_iccv17,Marszalek2009,non_local} largely focused on the \textit{classification} task (label each trimmed clip with an action label). For detection, where we want to assign different labels to different actors in the same scene, \emph{actor-centric} relationships need to be extracted. Training this in a fully supervised manner would require detailed labeling of actors and relevant objects~\cite{hico_det_wacv2018,gupta2015exploring}; such annotations can be very expensive to obtain. Therefore, we aim to build an action detection system that can infer actor-object spatio-temporal relations automatically with only actor-level supervision.

In this paper, we propose an action detection model that learns spatio-temporal relationships between actors and the scene. Motivated by the recent work of Santoro et al.~\cite{RN_deepmind17} on visual question answering, we use neural network to compute pair-wise relation information from the actor and scene features, which enables the module to be jointly trained with the action detector. We simplify the search space of scene features to be individual cells on a feature map, and pool the actor feature to be $1\times 1$. These simplifications allow us to compute relation information efficiently with $1\times 1$ convolutions. A set of $3\times 3$ convolutions are then used to accumulate relation information from neighboring locations. We refer to this approach as actor-centric relation network (ACRN). Finally, we also use the temporal context as inputs to ACRN. Such context is captured by 3D ConvNets as suggested by~\cite{ava_cvpr18}.

We evaluate our approach on JHMDB~\cite{jhmdb} and the recently released AVA dataset~\cite{ava_cvpr18}. Experimental results show that our approach consistently outperforms the baseline approach, which focuses on the actor, and alternative approaches that employ context information. We also visualize the relation heat maps with classification activation mapping~\cite{zhou2016cvpr_cam}. Figure~\ref{fig:teaser} shows two examples of such visualization. It is evident that ACRN learns to focus on the ball and its motion over time (flattened into 2D).

The primary contribution of this paper is to learn actor-centric spatio-temporal relationships for action detection in video. The rest of the paper describes our approach and experiments in detail. In section 2, we first review related work. In section 3, we present our approach to detect human actions. In section 4, we discuss several experiments on two datasets where we obtain state-of-the-art action detection performance.

%% file: related.tex
\section{Related work}

\noindent{\bf Action recognition.} Action recognition has traditionally focused on classifying actions in short video clips.  State-of-the-art methods rely either on two-stream 2D ConvNets~\cite{Karpathy2014,simonyan2014}, 2D ConvNets with LSTMs~\cite{lrcn2015,ng_cvpr2015} or 3D ConvNets~\cite{i3d_cvpr17,tran2015}.
While action classification in videos has been successful, it is inherently limited to short trimmed clips. If we want to address long untrimmed videos, temporal
localization is necessary in addition to action classification. This requires an additional step of determining the start and end time of each action instance. Many recent state-of-the-art methods~\cite{buch_cvpr17,dai_iccv17,xu_iccv17} rely on  temporal proposals and classification approaches similar in spirit to recent methods for object detection~\cite{ren2015faster}.

However, a more detailed understanding of actions in video requires localization not only in time, but also in space. This is particular true in the case of multiple actors~\cite{ava_cvpr18}. Many existing state-of-the-art approaches for spatio-temporal action localization~\cite{gkioxari2015,peng2016multi,saha2016,Singh_ICCV2017,weinzaepfel2015} employ state-of-the-art object detectors~\cite{SSD-Multibox,ren2015faster} to discriminate between action classes at the frame level. Recently, some approaches incorporate temporal context from multiple frames. This is particularly important for disambiguating actions such as ``stand up'' and ``sit down'', which may appear identical at the single frame level.
The tubelet approach~\cite{tubelets_iccv17} concatenates SSD features~\cite{SSD-Multibox} over spatio-temporal volumes and jointly estimates classification and regression over several frames. T-CNN~\cite{T_CNN_iccv17} uses 3D convolutions to estimate short tubes, micro-tubes rely on two successive frames~\cite{micro_tube2017} and pose-guided 3D convolutions add pose to a two-stream approach~\cite{pose_brox2017}.  Gu et al.~\cite{ava_cvpr18} rely on inflated 3D ConvNet (I3D) convolutions~\cite{i3d_cvpr17} for Faster R-CNN~\cite{ren2015faster} region proposals and show that the use of I3D over relatively long temporal windows~\cite{varol_pami17} improves the performance. The spatio-temporal separable 3D ConvNet (S3D)~\cite{s3dg_2017} improves the I3D architecture by observing that the 3D convolutions can be replaced by separable spatial and temporal convolutions without loss in accuracy, and that using such convolutions in higher layers of the network results in faster and more accurate models. We use the S3D~\cite{s3dg_2017} model as the baseline approach in this paper. 

Whereas several recent approaches for spatio-temporal action localization do take into account temporal information, they ignore spatial context such as interaction with humans, objects and the surrounding scene. This results in confusion of similar actions and interactions, such as jumping and shooting a basketball. 
We demonstrate that augmenting a state-of-the-art action localization approach with spatial context generates a significant performance improvement. 
~\\

\noindent{\bf Context in vision.} The use of context information to improve visual recognition has been extensively studied in computer vision. Early work showed that context can help scene classification~\cite{oliva2001modeling}, object detection~\cite{gupta2015exploring,heitz2008learning,mottaghi2014role,rabinovich2007objects,torralba2003context}, and action recognition in images~\cite{yao2010modeling}. In these cases, context often provides a strong prior that enables more robust recognition.  While these models of context were largely hand-designed, recent investigations have studied how to learn context with deep convolutional networks~\cite{shrivastava2016contextual,shrivastava2016beyond}. Spatial context has also been studied in self-supervised learning for learning unsupervised visual representations~\cite{doersch2015unsupervised,pathak2016context}. Beyond images, context has been leveraged in video, in particular for recognizing actions with hand-crafted features~\cite{Marszalek2009} and learned representations~\cite{girdhar2017attentional,sharma2015action}. While we are also interested in recognizing human actions with context, this paper focuses on the role of context for \emph{detection}. Importantly, since recognizing actions from crops is challenging even for humans, we believe context should play a critical role for learning robust action detection models.

Modeling the relations between objects~\cite{Peyre_2017_ICCV,lu2016visual} and more specifically between humans and objects~\cite{gkioxari2017interactnet,gupta2015} has been shown to improve the performance of recognizing relations in static images. Recent work~\cite{gkioxari2017interactnet} obtains state-of-the-art performance for  human-action-object recognition on V-COCO~\cite{gupta2015} and HICO-DET~\cite{hico_det_wacv2018}. In contrast to our approach, their model is only applied to static images and relies on full supervision of actor, action and objects as annotated in V-COCO~\cite{gupta2015} and HICO-DET~\cite{hico_det_wacv2018}.

%% file: approach.tex
\section{Action detection with actor-centric relation network}

\begin{figure}[t]
    \centering
    \includegraphics[width=0.95\linewidth]{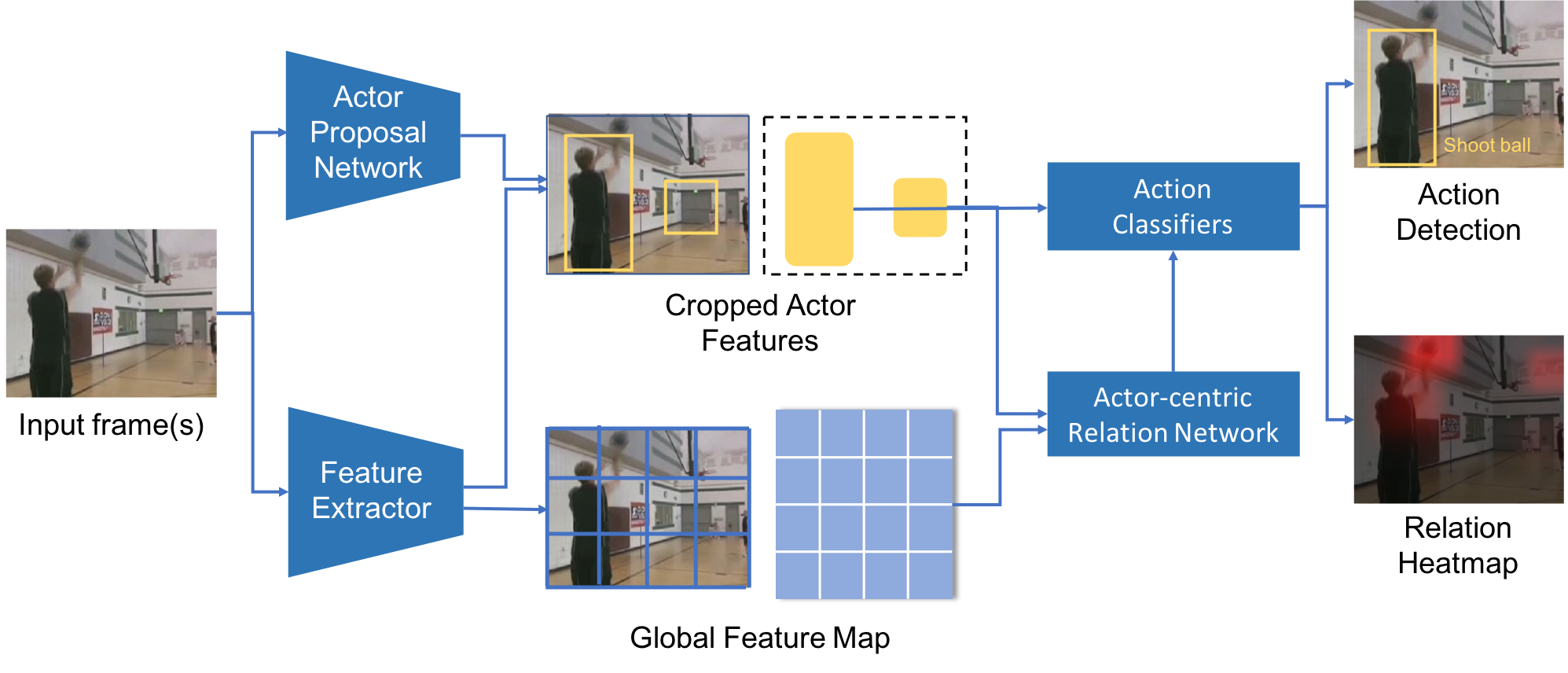}
    \caption{Overview of our proposed action detection framework. Compared to a standard action detection approach, the proposed framework extracts pairwise relations from cropped actor features and a global feature map with the actor-centric relation network (ACRN) module. These relation features are then used for action classification.}
    \label{fig:overall_flow}
\end{figure}

This section describes our proposed action detection framework. The framework builds upon the recent success of deep learning methods for object and action detection from static images~\cite{ren2015faster} and videos~\cite{peng2016multi}. We note that the relational information between the actor of interest and other actors or objects are important to identify actions, but are typically ignored by recent action detection methods~\cite{peng2016multi,tubelets_iccv17}; such annotations could be time consuming to collect, and are not provided by many of the recent action recognition datasets~\cite{kinetics17,ava_cvpr18,charades2016}. Our proposed frameworks aims at explicitly modeling relations with weak actor-level supervision, with an actor-centric relation network module. Once trained, the framework can not only detects human actions with higher accuracy, but can also generate spatial heat maps of the relevant relations for each actor and action. An overview of the approach can be found in Figure~\ref{fig:overall_flow}.

\subsection{Action detection framework}\label{sec:framework}
Our goal is to localize actions in videos. We follow the popular paradigm of frame-based action detection, where the model produces bounding-box predictions for actions on each frame individually~\cite{peng2016multi,gkioxari2015}, and then links them into tubes as a post-processing step.

\medskip\noindent\textbf{Action detection model.} Our base model has two key components: actor localization and action classification. These two components are trained jointly in an end-to-end fashion. This action detection model was proposed in~\cite{ava_cvpr18}, motivated by the success of applying end-to-end object detection algorithms to action detection~\cite{peng2016multi,tubelets_iccv17}. The inputs to the base model include the key frame to generate action predictions, and optionally neighboring frames of the key frame as temporal context. The outputs of the base model include 2D bounding boxes of localized actions for the key frame. The overall architecture of the base model largely resembles the Faster R-CNN detection algorithm. For actor localization, our method uses the region proposal network (RPN) from Faster R-CNN to generate 2D actor proposals. For action classification, we use deep representations extracted from the key frame and (optionally) neighboring frames. Unlike Faster R-CNN, we do not require the actor proposal network to share the same extracted features as the action classification network, although such sharing is possible. This allows more freedom to the choice of action classification features without adding much computation overhead, as classification feature computation is usually dominated by the neighboring frames.

\medskip\noindent\textbf{Incorporating temporal context.} We adopt 3D ConvNets~\cite{i3d_cvpr17,s3dg_2017} as used by~\cite{ava_cvpr18} to incorporate larger temporal context from neighboring frames. We found that 3D ConvNets consistently outperform alternative approaches such as channel-wise stacking of frames at the input layer or average pooling at the output layer. The output feature map from 3D ConvNets has an extra time dimension, which is inconsistent with the 2D bounding box proposals generated by RPN. We address this issue by \textit{flattening} the 3D feature map with a $\texttt{t}\times 1\times 1$ temporal convolution, where $\texttt{t}$ is the size of the time dimension. The flattened 2D feature map can then be provided to a standard differentiable ROIPooling operation~\cite{STN,huang2016coco} to produce cropped actor features. The cropped actor features are then inflated back to 3D, to allow reusing the pre-trained 3D ConvNets weights for classification. Empirically, we find that the flattening approach gives on par or better accuracy than keeping the temporal dimension with 3D ROIPooling.

\medskip\noindent\textbf{Architecture details.} For accurate actor bounding-box locations, we follow~\cite{ava_cvpr18} and use a 2D ResNet-50 model~\cite{he2016resnet} trained on key frames with action bounding-box annotations. For action classification, we use gated separable 3D network (S3D-G)~\cite{s3dg_2017}. Compared with I3D~\cite{i3d_cvpr17} used in~\cite{ava_cvpr18}, S3D-G replaces full 3D convolutions with separable spatial and temporal convolutions, and employs spatio-temporal feature gating layers. Overall, S3D-G is faster, has fewer parameters, provides higher accuracy compared to other 3D ConvNet models, and has a flexible design which makes it ideal for the large-scale action detection setup.

Following the recommendation from~\cite{s3dg_2017}, we use the \textit{top-heavy} configuration and use 2D convolutions without gating until the \mixed{4b} block (we follow the same naming conventions as the Inception networks~\cite{inception}), and switch to separable 3D convolutions with gating onwards. To combine RGB and optical flow input modalities, we use early fusion at the \mixed{4f} block instead of late fusion at the logits layer. With these changes, we observed a $1.8\times$ speed-up in our action detection model without loosing performance. We use the features from the fused \mixed{4f} ($\texttt{t}\times\texttt{h}\times\texttt{w}\times\texttt{c}$) block for action classification. These features have a spatial output stride of $16$ pixels and a temporal output stride of $4$ frames. Regions in \mixed{4f} corresponding to actor RPN proposals are temporally flattened and used as the input for the action classification network. We will refer to the $\texttt{h}\times\texttt{w}\times\texttt{c}$ feature map as $\mathcal{F}$ going forward. For each RPN proposal generated by a potential actor ($b_i = (x^i_1, y^i_1, x^i_2, y^i_2)$)), we crop and resize the feature within $b_i$ from $\mathcal{F}$ using ROIPooling to obtain a fixed-length representation $\mathcal{F}\left(b_i\right)$ of size $7\times7\times832$. This feature representation is used by the action classifier, which consists of \mixed{5b} and \mixed{5c} blocks (that output $7\times7\times1024$ feature), and an average pooling layer which outputs $1\times1\times1024$ feature. This feature is then used to learn a linear classifier for actions and a regressor for bounding-box offsets. We refer to the action detection model described above as our \textbf{Base-Model} throughout this paper. As shown in the experiments, the Base-Model by itself obtains state-of-the-art performance for action detection on the datasets explored in this work.

\subsection{Actor-centric relations for action detection}
A key component missing in the Base-Model is reasoning about relations outside the cropped regions. It is important to model such relations, as actions are in many cases defined by them (see Figure~\ref{fig:teaser}). 
Here, we propose to extract actor-centric relation features and to input them to the action classification network. Our approach performs relation reasoning given only action annotations, and  automatically retrieves the regions that are most related to the action. Thus, we refer to our approach as an actor-centric relation network (ACRN).

\medskip\noindent\textbf{Actor-centric relations.} Given an input example $I$ with a set of actors $\mathcal{A} = \{A_1,A_2, ..., A_M\}$ and objects $\mathcal{O} = \{O_1,O_2, ..., O_N\}$, we define a pair-wise relation feature between actor $A_i$ and object $O_j$ as 
$g_\theta(a_i,o_j)$, where $g_\theta(\cdot)$ is a feature extractor function parameterized by $\theta$, and $a_i$ and $o_j$ are the feature representations of actor $A_i$ and object $O_j$ respectively.

The actor-centric relation feature for actor $A_i$ can be computed by
\begin{equation}
\text{ACR}(A_i) = f_\phi \bigg(\Big\{g_\theta(a_i,o_j): O_j \in \mathcal{O} \Big\} \bigg) ,
\end{equation}
where $f_\phi(\cdot)$ is a function that aggregates features from all pair-wise relations, parameterized by $\phi$.

To use actor-centric relations for action detection, we need to define actors, objects and their relations. We treat each actor proposal generated by RPN as one actor. However, objects and their relations are not explicitly defined or annotated for the action detection task. We adopt a workaround that treats each individual feature cell in the convolutional feature map $\mathcal{F}$ as an \textit{object} $O_i$, which naturally gives the object representation $o_i$. This simplification avoids the need of generating object proposals, and has been shown to be effective for video classification~\cite{non_local} and question answering~\cite{RN_deepmind17} tasks. However, as we show in the experiment section, directly applying this simplification does not improve the performance of the Base-Model. We compute $\text{ACR}(A_i)$ with neural networks, this allows the module to be end-to-end trained with the Base-Model. In particular, both $g_\theta$ and $f_\phi$ can be implemented with standard convolutions and pooling operations.

\medskip\noindent\textbf{Action detection with actor-centric relation network (ACRN).} We now discuss how to incorporate ACRN into our action detection framework. Given $N$ frames from a video ($\mathcal{V}$), we first extract the fused feature map $\mathcal{F}_v$ of size $\texttt{h}\times\texttt{w}\times\texttt{c}$ and a set of bounding-boxes generated by the actor RPN ($\mathcal{B} = (b_i, ..., b_R)$). For each box $b_i$, we follow the procedure described in Section~\ref{sec:framework} to get an actor feature $f^a_i$ of size $1\times1\times1024$. Note that this is the same feature used by the Base-Model for action classification.

To extract pair-wise relation features, we follow the relation network module used by Santoro et al.~\cite{RN_deepmind17} and implement $g_\theta$ as a single fully-connected layer. The inputs to $g_\theta$ are set as the concatenation of features from one actor proposal and one object location, along with their locations:

\begin{equation}
    a_i = \left[f^a_i; b_i\right] \text{, and } o_{j,k} = \left[\mathcal{F}_v(j,k); l\left(j,k\right)\right], 
\end{equation}
where $\mathcal{F}_v(j,k)$ is the $1\times1\times832$ feature extracted at feature location $(j,k)$, and $l = (j/H, k/W)$.

\begin{figure}[t]
    \centering
    \includegraphics[width=0.95\linewidth]{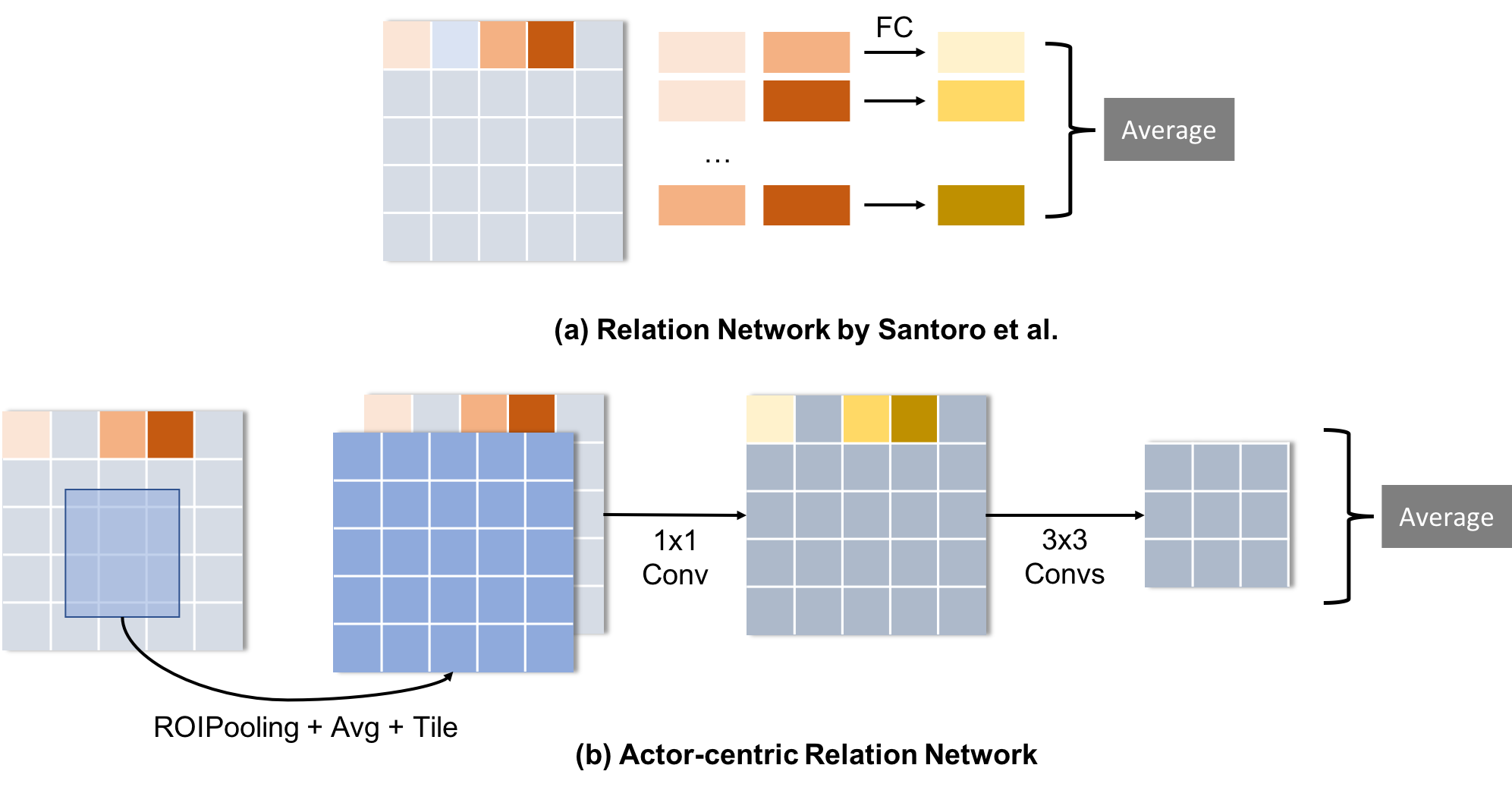}
    \caption{Comparison of our proposed ACRN (\textbf{b})  with the relation network used by Santoro et al.~\cite{RN_deepmind17} (\textbf{a}). We compute relation feature maps by duplicating actor features and applying a 1$\times$1 convolution. A set of 3$\times$3 convolutions are then applied on relation feature map to accumulate information from neighboring relations.}
    \label{fig:acrn_compare}
\end{figure}

In practice, we can efficiently compute $g_\theta(a_i, o_{j,k})$ for all $(j,k)$ locations using convolution operations. The actor appearance feature $f^a_i$ is duplicated to $h\times w\times1024$ feature and concatenated with $\mathcal{F}$ channel-wise, along with the box and location embeddings. Next, $g_\theta$ is computed using a $1\times1$ convolution layer, which outputs $\mathcal{F}^\theta(a_i)$ of size $\texttt{h}\times \texttt{w}\times832$. These operations are illustrated in Figure~\ref{fig:acrn_compare} (b). Since $a_i$ and $o_{j,k}$ come from different layers with varying feature amplitude, we follow the design decisions from~\cite{bell2016inside,parsenet,shrivastava2016contextual} to normalize and scale the features when combining them as inputs to the relational reasoning modules.

After $\mathcal{F}^\theta(a_i)$ is computed, the model needs to aggregate all $g_\theta$ with $f_\phi$. One option is to directly apply average pooling as in~\cite{RN_deepmind17}, which works for synthetic data with relative simple scene~\cite{CLEVR}. However, for action recognition the relevant relational information could be very sparse, and averaging will dilute such information. Moreover, since the relation features are computed locally, information about bigger objects could be lost after average pooling. Instead, we propose to apply convolution operations on $\mathcal{F}^\theta(a_i)$ before average pooling, which allows relational information to accumulate over neighboring locations. In practice, we use \mixed{5b} and \mixed{5c} blocks of the S3D-G network and an average pooling layer (similar to the action classifier network) to output a $1\times1\times1024$ feature ($f^{\text{RN}}_i$). Finally, we concatenate actor feature $f^a_i$ and spatio-temporal relation feature $f^{\text{RN}}_i$ to get a $1\times1\times2048$ representation which is used for action classification and bounding-box regression for a given actor box $b_i$ (see Figure~\ref{fig:overall_flow}). The same process is repeated for all proposed regions $\mathcal{B}$.

To handle objects of various scales, we further extend the ACRN module to allow feature maps other than $\mathcal{F}$ to be used, each of which can be resized to different scales. For example, smaller objects might be better represented in the earlier layers of the network at a larger scale feature map, while bigger objects might be better represented in the higher layers at a smaller scale. In Section~\ref{sec:ablation}, we study the impact of using different layers and their combination for relation reasoning.

%% file: experiment.tex
\section{Experiments}
In this section, we perform design and ablation analysis of our method and visualize what relations are captured by the actor-centric relation network. Finally, we evaluate it on the task of spatio-temporal action localization and demonstrate consistent and significant gain across multiple benchmarks.

\subsection{Experimental Setup}
\medskip\noindent\textbf{Datasets and metrics.} We report results on the JHMDB~\cite{jhmdb} and AVA~\cite{ava_cvpr18} action detection benchmarks. JHMDB~\cite{jhmdb} consists of 928 temporally trimmed clips with 21 action classes and has three training/validation splits. Unless noted otherwise, we follow the standard setup and report results by averaging over all three splits. We report the frame-level and video-level mean average precision (frame-AP and video-AP) with an intersection-over-union (IOU) threshold of 0.5. For video-AP, we link per-frame action detection results into tubes using the algorithm from~\cite{ava_cvpr18}. We use the AVA version 2.1 benchmark~\cite{ava_cvpr18}. It consists of 211k training and 57k validation examples labeled at 1FPS over 80 action classes. We follow their baseline setup and report results on 60 action classes which have at least 25 validation examples per class. We report frame-AP for AVA.

\medskip\noindent\textbf{Implementation details.} 
For the Base-Model, we use the ResNet-50~\cite{he2016resnet} RGB model for actor localization and the S3D-G~\cite{s3dg_2017} two-stream model for action classification. The detailed architecture is described in Section~\ref{sec:framework}. Optical flow for the two-stream network is extracted using FlowNet2~\cite{flownet2}. As is standard practice, the ResNet-50 model is pre-trained on ImageNet and S3D-G RGB+Flow streams are pre-trained on Kinetics. The classification head (\mixed{5b}, \mixed{5c}) are initialized from RGB stream pre-trained on Kinetics for both RN and actor classification, but they are updated separately (weights are not shared). The whole pipeline (actor localization, actor-centric RN, and action classification) is trained jointly end-to-end. 

We train the model for 200K and 1.2M steps for JHMDB and AVA respectively, and use start asynchronous SGD with a batch-size of 1 per GPU (11 GPUs in total), mini-batch size of 256 for actor RPN and 64 for action classifier (following~\cite{ava_cvpr18}). We warm-start the learning rate from 0.00001 to 0.001 in 40K steps using linear annealing and then use cosine learning rate decay~\cite{cosine_lr}. To stabilize training, the batch-norm updates are disabled during training and we apply a gradient multiplier of 0.01 to gradients from RN to the feature map.

\subsection{Design and ablation analysis}
\label{sec:ablation}
We perform a number of ablation experiments to better understand the properties of our actor-centric relation network and its impact on the action detection performance. The results are shown in Table~\ref{tab:ablation_jhmdb} and Table~\ref{tab:feature_layer_ava}.

\begin{table}[t]
\begin{center}
\subfloat[Relation reasoning modules.\label{tab:ablation_instantiation}]{
\begin{tabular}{c|c}
\hline
Model  & frame-AP \\
\hline
Base-Model & 75.2 \\
\hline
Resize+Concat~\cite{bell2016inside,parsenet,shrivastava2016contextual} & 74.8 \\
Santoro et al.~\cite{RN_deepmind17} & 75.1 \\
ACRN& \textbf{77.6} \\
\hline
\end{tabular}
}
\qquad
\subfloat[Temporal context.\label{tab:ablation_time}]{
\begin{tabular}{c|c|c}
\hline
Frames  & Base-Model & ACRN \\
\hline
1 & 52.6 & 54.0 \\
5 & 66.1 & 69.8 \\
10 & 70.6 & 74.9 \\
20 & 75.2 & 77.6 \\
\hline
\end{tabular}
}
\\
\subfloat[Feature layers and scales.\label{tab:ablation_feature_layer}]{
\begin{tabular}{c|c|c|c|c|c|c}
\hline
Feature & \texttt{Conv1a} & \texttt{Conv2c} & \mixed{3b} & \mixed{4b} & \mixed{4f} & \texttt{Conv2c}, \mixed{3c}, \mixed{4f}\\
\hline
Scale 0.5 & 76.4 & \textbf{77.2} & 76.2 & 76.2 & 77.1 & \textbf{77.9} \\
Scale 1.0 & 76.6 & \textbf{77.9} & 76.4 & 76.6 & 77.6 & 77.5 \\
\hline
\end{tabular}
}
\end{center}
\caption{Frame-AP evaluating the impact of different parameters of ACRN on JHMDB dataset (3 splits).}
\label{tab:ablation_jhmdb}
\end{table}

\medskip\noindent\textbf{Importance of relation reasoning modules.} Table~\ref{tab:ablation_instantiation} compare the performance between the Base-Model which only uses actor features, and three different relation reasoning modules which take global feature maps as additional inputs. We L2-normalize the actor appearance feature $f^a_i$ and scene context feature \mixed{4f}, concatenate the features together, scale the L2-norm of the concatenated feature back to the L2-norm of $f^a_i$, and use a $1\times1$ convolution layer to reduce the number of channels to be same as $f^a_i$. We study the following relational reasoning modules:

\textbf{Resize+Concat}~\cite{bell2016inside,parsenet,shrivastava2016contextual}: resize the global feature map and actor feature map at \mixed{4f} to have same size and directly concatenate channel-wise. The concatenated feature maps are fed into the classification head (\mixed{5b-5c}).

\textbf{Santoro et al.}~\cite{RN_deepmind17}: global and actor feature maps are used to compute $g_\theta$, which are then averaged to compute $f_\phi$ with one fully-connected layer.

\textbf{ACRN}: global and actor feature maps are used to compute relation feature maps by ACRN, which are fed into the classification head.

Table~\ref{tab:ablation_instantiation} shows performance comparisons in frame-AP on JHMDB. Our proposed ACRN imrpoves by \textbf{2.4} over the Base-Model. However, Resize+Concat and Santoro et al. fail to outperform the baseline despite having access to global feature maps. The gap highlights the importance of designing an appropriate relation reasoning module.

\begin{table}
    \centering
\subfloat{
\begin{tabular}{c|c|c|c}
\hline
Action & \mixed{4f} & \texttt{Conv2c} & Gap\\
\hline
answer phone & 56.0 & 50.3 & 5.7 \\
jump & 6.8 & 4.0 & 2.8 \\
swim & 35.1 & 33.2 & 1.9 \\
read & 10.3 & 8.6 & 1.7 \\
dance & 32.7 & 31.6 & 1.1 \\
\hline
\end{tabular}
}
\qquad
\subfloat{
\begin{tabular}{c|c|c|c}
\hline
Action & \mixed{4f} & \texttt{Conv2c} & Gap\\
\hline
drive & 15.3 & 19.5 & -4.2 \\
fight & 36.2 & 40.4 & -4.2 \\
kiss & 15.6 & 19.4 & -3.8 \\
play instrument & 7.1 & 10.9 & -3.8 \\
touch & 24.2 & 26.7 & -2.5 \\
\hline
\end{tabular}
}
\caption{AVA actions with biggest performance gaps when different features are used by ACRN.}
\label{tab:feature_layer_ava}
\end{table}

\medskip\noindent\textbf{Impact of temporal context.}
Table~\ref{tab:ablation_time} studies the impact of temporal context on the Base-Model and the proposed ACRN framework by varying the number input frames, and show the results in . As observed by~\cite{ava_cvpr18}, using more input frames generally helps our models. ACRN consistently improves over the Base-Model across all temporal lengths.

\medskip\noindent\textbf{Comparison of feature layers and scales.} ACRN can take feature maps from the different layers of a ConvNet as inputs. Each feature map can be resized to have different scales. Choices of feature layer and scale may have pros and cons: intuitively, features from higher layers (e.g. \mixed{4f}) encode more semantic information, but have lower resolution; features from lower layers (e.g. \texttt{Conv2c}) have higher resolution but are less semantically meaningful. Similarly, feature map with larger scale potentially helps to identify interactions involving smaller objects, but also increases the number of relations to aggregate.

In Table~\ref{tab:ablation_feature_layer}, we report frame-mAP on the JHMDB dataset by varying the feature layers and scales of the global feature map. We observe that \texttt{Conv2c} is the best performing single feature, followed by \mixed{4f}. The performance is relatively stable for different scales. We note that combining features from multiple layers not necessarily results in better overall performance. In Table~\ref{tab:feature_layer_ava}, we list the AVA categories with highest performance gap when using \mixed{4f} and \texttt{Conv2c}. We can see that the two feature layers are clearly complimentary for many actions. In the following experiments, we report ACRN results based on the best single feature layer and scale.

\subsection{Comparison with the state of the art}
We compare our best models with the state-of-the-art methods on JHMDB and AVA. For the state-of-the-art methods, we use the same experimental setup and quote the results as reported by the authors. We fix the number of input frames to 20 for I3D, Base-Model and ACRN.

\begin{figure}[t]
    \centering
    \includegraphics[width=.95\linewidth]{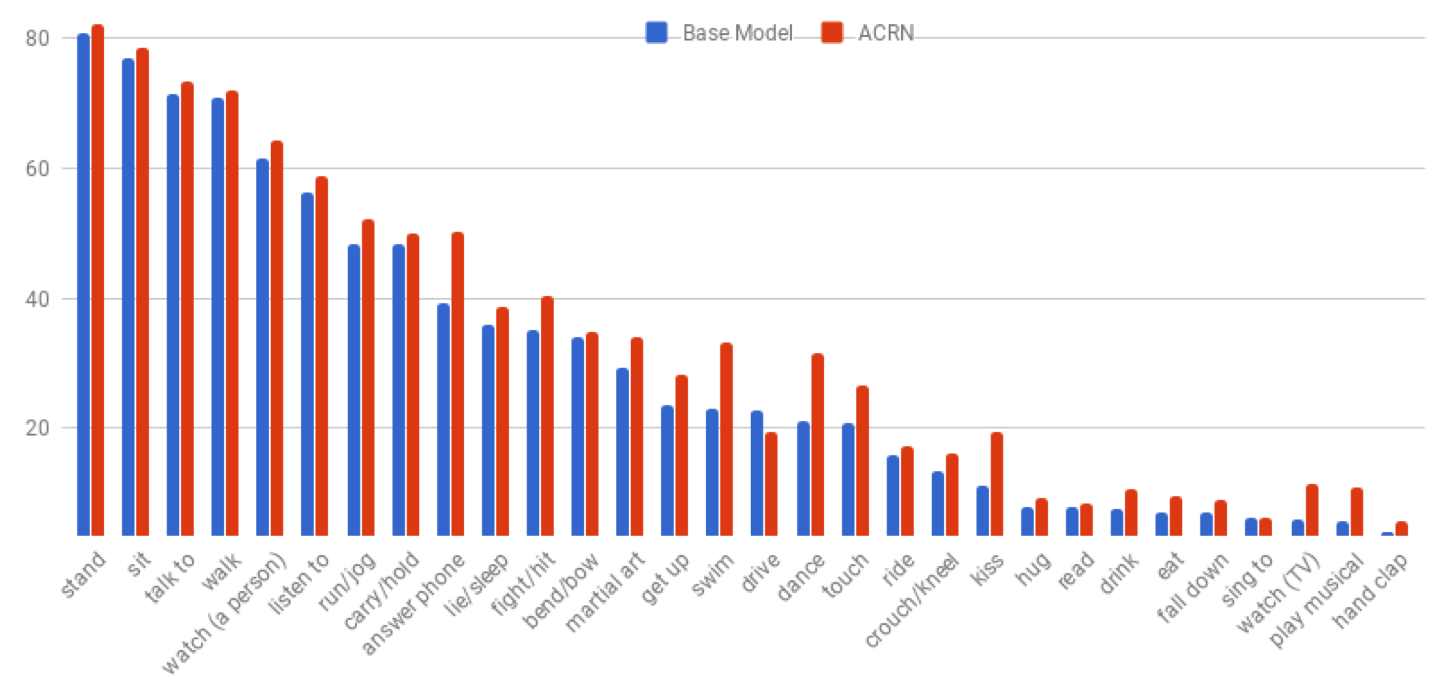}
    \caption{Per-category frame-AP comparison between the Base-Model and ACRN on AVA. Labels are sorted by descending Base-Model performance, only the top 30 categories are shown.}
    \label{fig:ava2_ap}
\end{figure}

As shown in Table~\ref{tab:rn_sota}, our Base-Model already outperforms all previous methods, and the proposed ACRN algorithm further achieves a gain over this Base-Model. We also look into the per-class performance breakdown: on the JHMDB dataset, ACRN outperforms the Base-Model significantly for catch (12\%), jump (6\%), shoot gun (5\%) and wave (10\%). The gain is smaller when the performance of the Base-Model is almost saturated (e.g. golf, pullup and pour). The Base-Model performs only slightly better on pick, throw and run. When visualizing the relation heatmaps, we can see that ACRN has difficulty attending to the right relations for these actions.

On the AVA dataset, the per-class performance breakdown for the 30 highest performing categories can be found in Figure~\ref{fig:ava2_ap}. We can discover that the biggest gains are achieved for answer phone (11\%), fight (5\%), swim (10\%), dance (10\%), touch (6\%), kiss (8\%) and play musical instruments (5\%), most of which involve human-human or human-object interactions.

\begin{table}
\begin{center}
\subfloat[JHMDB (3 splits)]{
\begin{tabular}{c|c|c}
\hline
Model  & frame-AP & video-AP \\
\hline
Peng et al.~\cite{peng2016multi} & 58.5 & 73.1 \\
%Singh et al.~\cite{Singh_ICCV2017} & - & 72.0 \\
ACT~\cite{tubelets_iccv17} & 65.7 & 73.7 \\
I3D~\cite{ava_cvpr18} & 73.3 & 78.6 \\
Base-Model & 75.2 & 78.8 \\
ACRN & {\bf 77.9} & {\bf 80.1} \\
\hline
\end{tabular}
}
\qquad
\subfloat[AVA (version 2.1)]{\begin{tabular}{c|c}
\hline
Model  & frame-AP \\
\hline
Single frame~\cite{ava_cvpr18} & 14.2 \\
I3D~\cite{ava_cvpr18} & 15.1 \\
Base-Model & 15.5 \\
ACRN & \textbf{17.4} \\
\hline
\end{tabular}
}
\end{center}
\caption{Comparison with state of the art on (a) the JHMDB dataset and (b) AVA . For JHMDB, we report average precision over 3 splits.}
\label{tab:rn_sota}
\end{table}

\begin{figure}
\centering
\includegraphics[width=0.85\linewidth]{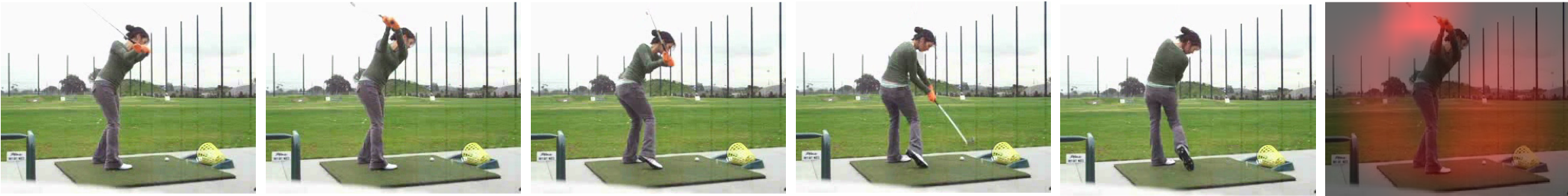}
\\
golf
\\
\includegraphics[width=0.85\linewidth]{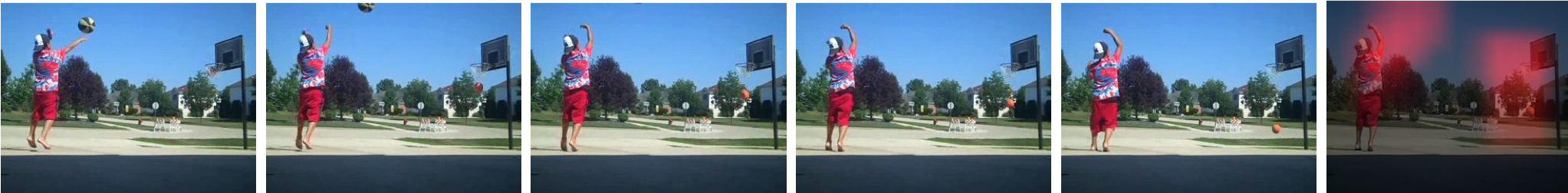}
\\
shoot ball
\\
\includegraphics[width=0.85\linewidth]{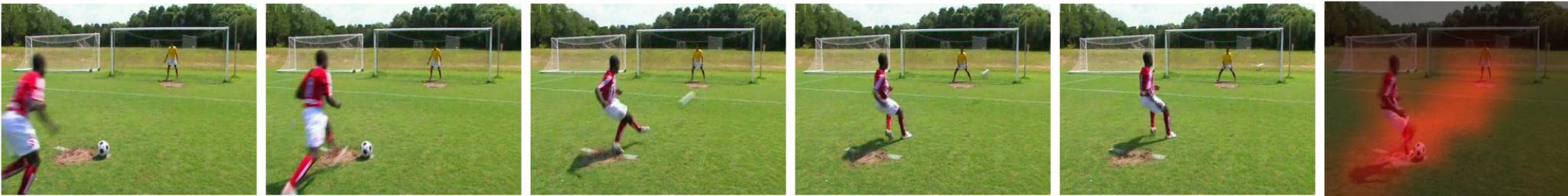}
\\
kick ball
\\
\includegraphics[width=0.85\linewidth]{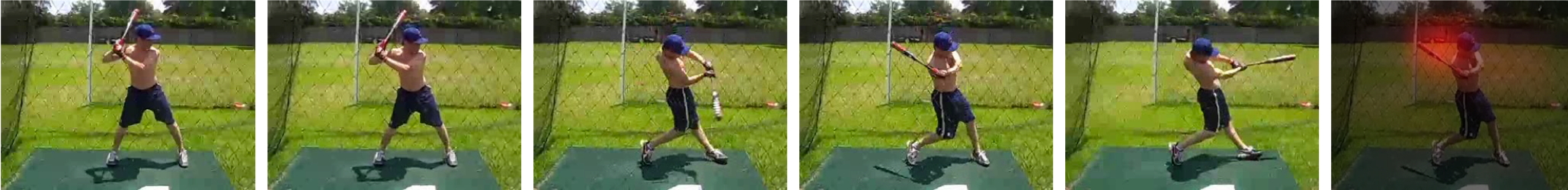}
\\
baseball swing
\\
\includegraphics[width=0.85\linewidth]{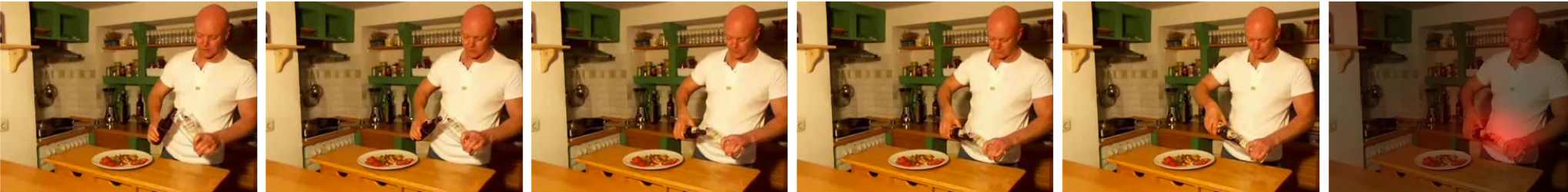}
\\
pour
\\
\caption{Visualization of relation heatmap on JHMDB dataset.}
\label{fig:viz_cam}
\end{figure}

\begin{figure}
\begin{center}
\begin{tabular}{cc}
\includegraphics[width=0.48\linewidth]{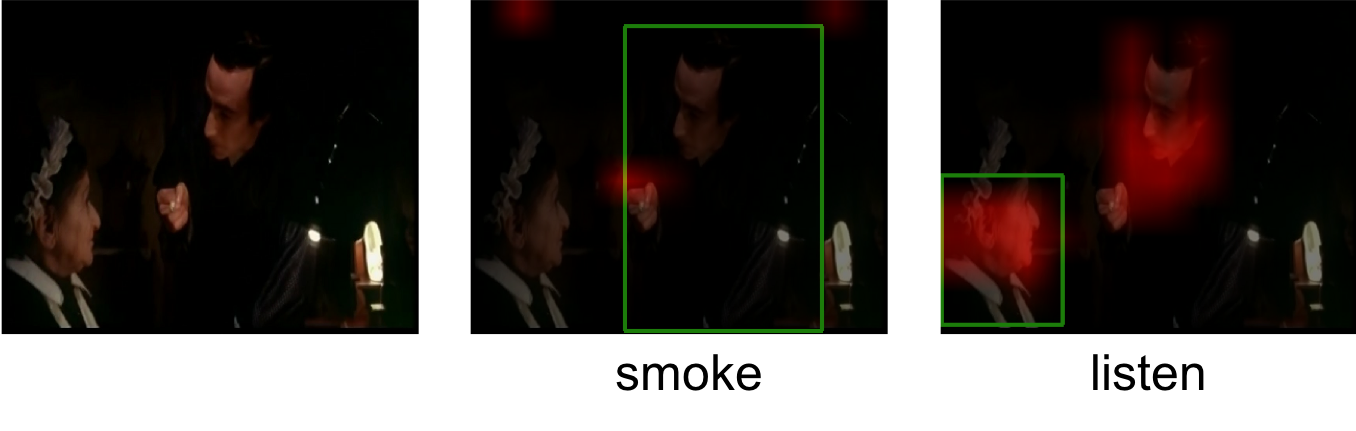}&
\includegraphics[width=0.48\linewidth]{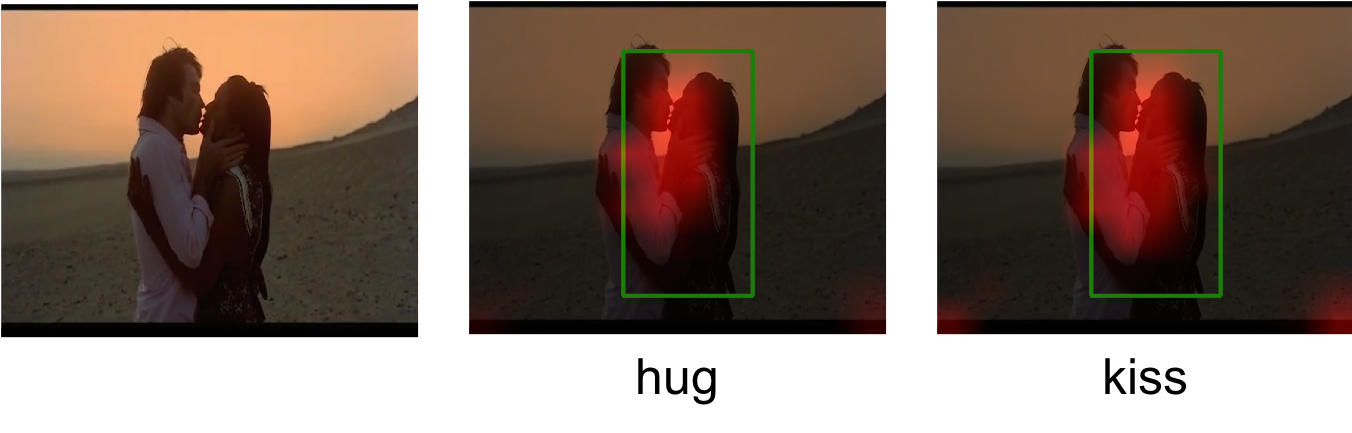}\\
\includegraphics[width=0.48\linewidth]{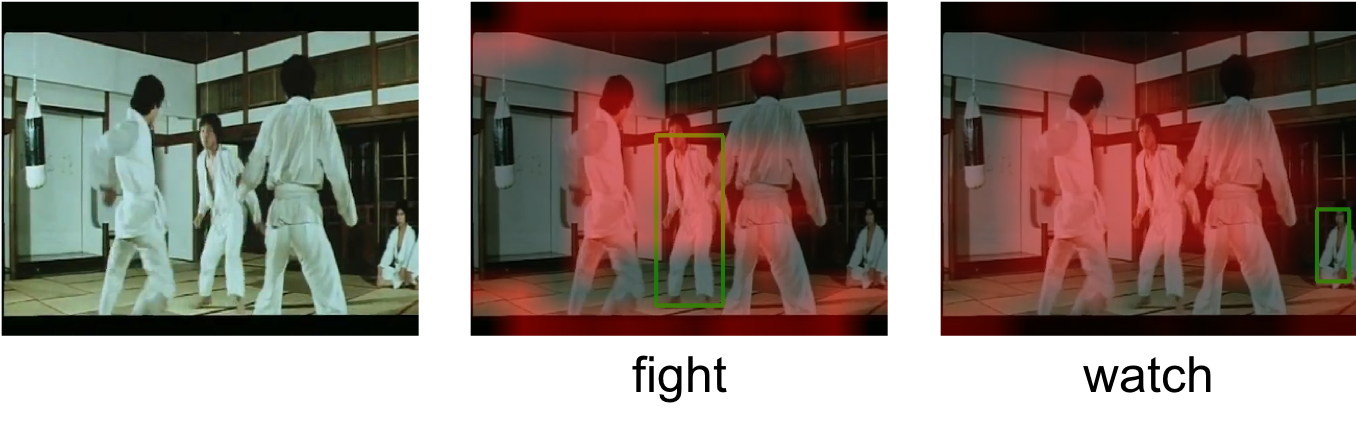}&
\includegraphics[width=0.48\linewidth]{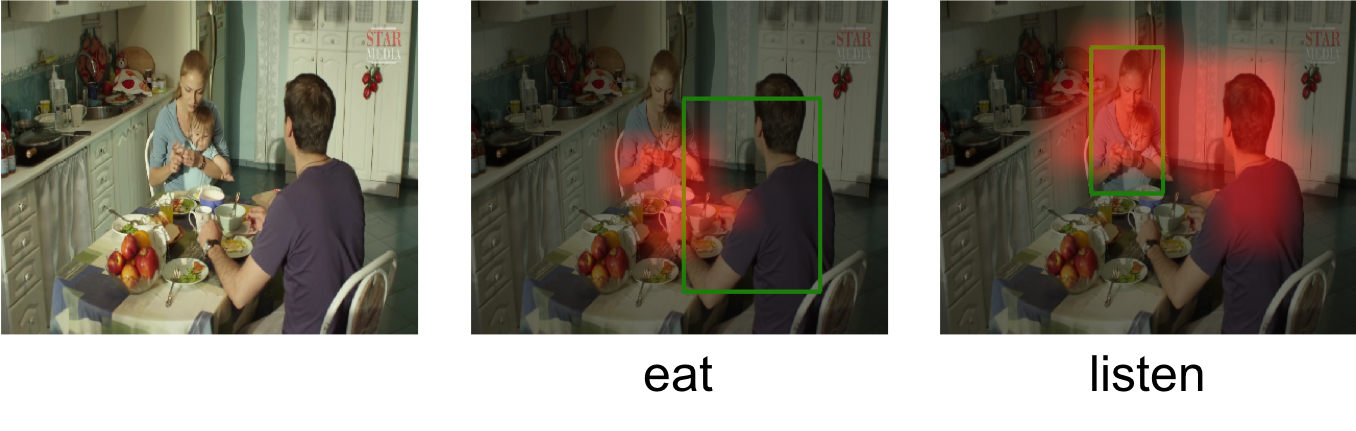}\\
\includegraphics[width=0.48\linewidth]{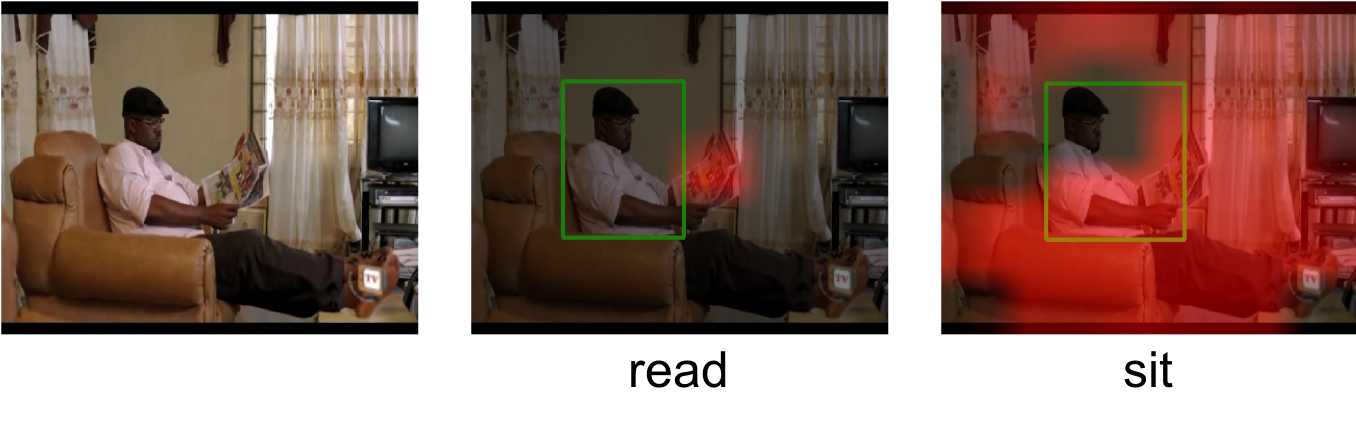}&
\includegraphics[width=0.48\linewidth]{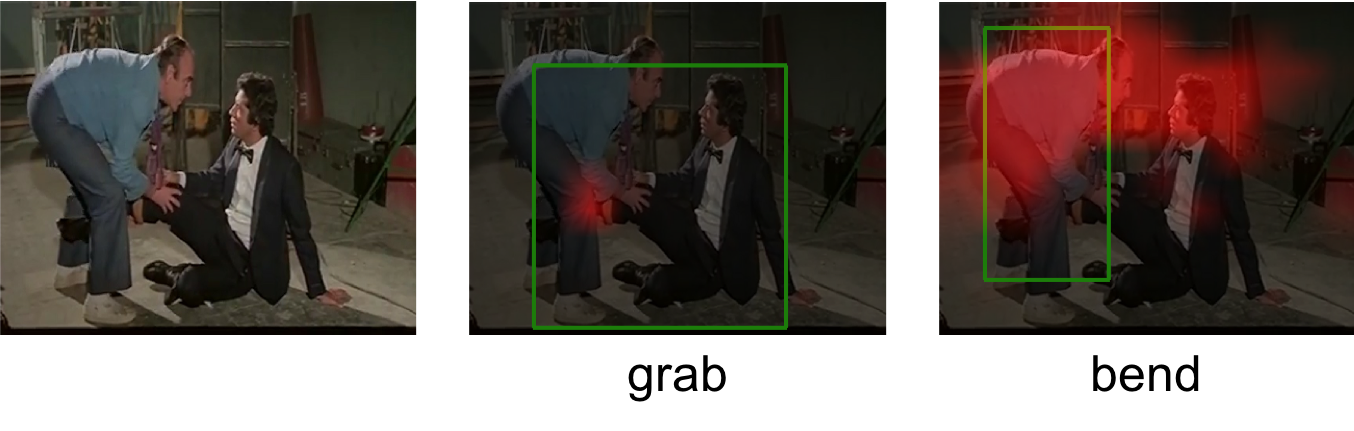}\\
\end{tabular}
\end{center}
\caption{Visualization of relation heatmap on AVA. The actor corresponding to the heatmap is marked in green, and its action is shown below the example. Notice how the heatmap varies depending on the action categories.}
\label{fig:ava_viz}
\end{figure}

\subsection{Qualitative Results}

To qualitatively verify what relations are learned by ACRN, we apply the class activation map (CAM)~\cite{zhou2016cvpr_cam} method to visualize the per-category relation heatmaps based on ACRN outputs. We modify the inference network by removing the average pooling operation after the ACRN branch, and apply the final action classifier as $1\times 1$ convolutions on the relation feature map. This allows us to generate spatially localized per-category activations, which illustrates the relations important to identify a certain action. Note that the spatial heatmaps also encode temporal information, as the input features are flattened from 3D to 2D.

\begin{figure}[t]
\centering
\includegraphics[width=0.95\linewidth]{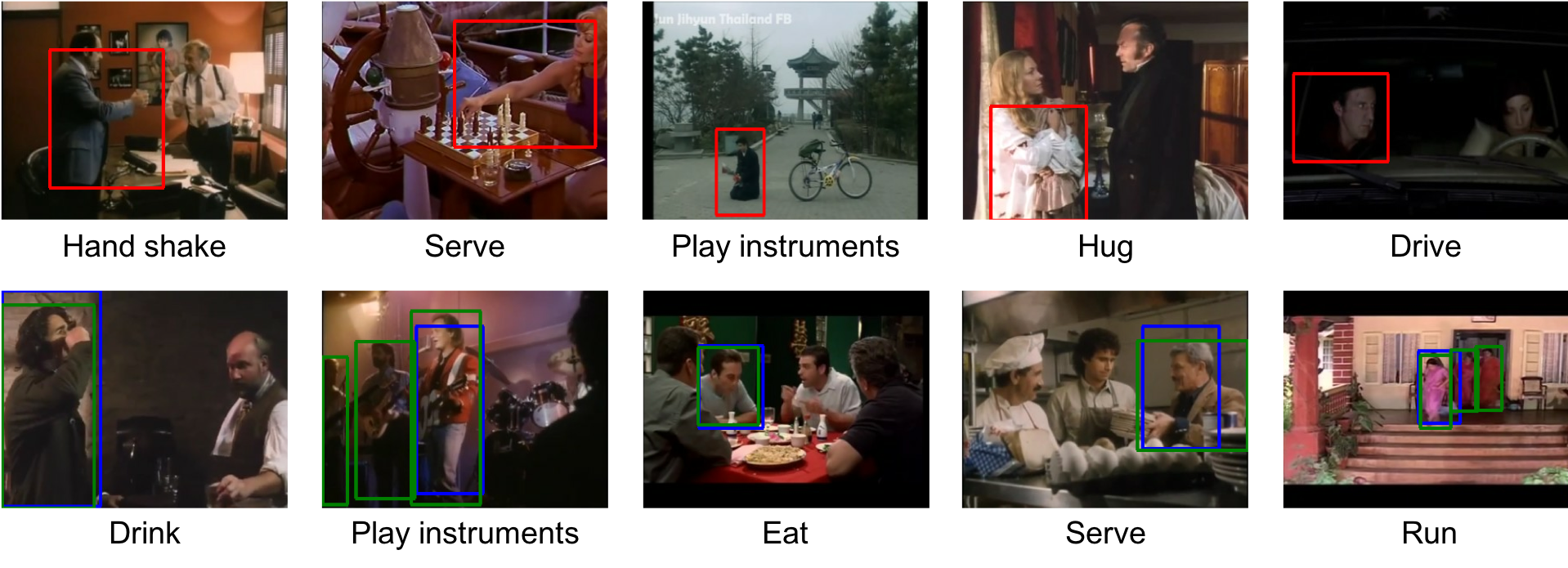}
\caption{(\textit{Top row}) False alarm detections from the Base-Model (red boxes) that are removed by ACRN. (\textit{Bottom row}) Miss detections (green) of the Base-Model captured by ACRN (blue).}
\label{fig:before_after}
\end{figure}

Figure~\ref{fig:viz_cam} and Figure~\ref{fig:ava_viz} show the visualizations of the top-1 and top-2 highest scoring detections on JHMDB and AVA respectively. We render the bounding boxes and their associated relation heatmap in green and red respectively. We can see that ACRN is able to capture spatio-temporal relations beyond the actor bounding box, and its output depends on actor and action. Finally, Figure~\ref{fig:before_after} illustrates examples for which the false alarms of the Base-Model are removed by ACRN (top row) and the missing detections are captured by ACRN (bottom row).

